%% file: main.tex
\documentclass[10pt,twocolumn,letterpaper]{article}
\usepackage[T1]{fontenc}
\usepackage[pagenumbers]{cvpr}
\definecolor{cvprblue}{rgb}{0.21,0.49,0.74}
\usepackage[breaklinks,colorlinks,allcolors=cvprblue]{hyperref}
\hypersetup{pdftitle={Reading Right, Answering Wrong: How Visual Configuration Changes Affect Evidence Use in VLMs},pdfauthor={Dingyang Lin, Yingfeng Luo, Chenglong Wang, Chenwei Zhu, Anxiang Ma, Jingbo Zhu, Tong Xiao},pdfsubject={Preprint}}
\graphicspath{{figures/}}
\newenvironment{keywords}{\par\noindent\textbf{Keywords: }\ignorespaces}{\par}
\title{Reading Right, Answering Wrong: How Visual Configuration Changes Affect Evidence Use in VLMs}
\input{authors}
\begin{document}
\ifdefined\pdfminorversion\pdfminorversion=7\else\fi
\raggedbottom
\maketitle
\input{sections/abstract}
\input{sections/keywords}

\input{sections/teaser_float}
\input{sections/introduction}
\input{sections/related_work}

\input{sections/boundary_table_float}
\input{sections/methodology}

\input{sections/results_and_analysis}

\input{sections/conclusion}
{\fontsize{9}{9.5}\selectfont
\setlength{\bibsep}{0pt}
\bibliographystyle{IEEEbib_dblp}
\bibliography{references}
}
\input{sections/ethics}
\end{document}

%% file: authors.tex
\author{
Dingyang Lin$^{1}$, Yingfeng Luo$^{1}$, Chenglong Wang$^{1}$, Chenwei Zhu$^{1}$,\\
Anxiang Ma$^{1}$, Jingbo Zhu$^{1,2}$, Tong Xiao$^{1,2}$\thanks{Corresponding author.}\\[4pt]
{\normalsize $^{1}$School of Computer Science and Engineering, Northeastern University, Shenyang, China}\\
{\normalsize $^{2}$NiuTrans Research, Shenyang, China}\\
{\tt\small ldy380641245@gmail.com\quad xiaotong@mail.neu.edu.cn}
}

%% file: sections/abstract.tex
\begin{abstract}
Vision--language models (VLMs) have achieved strong performance on tasks such as visual question answering, yet small image resizes can turn correct answers into errors. We investigate whether changes in visual configuration, such as image tiling and token arrangement, contribute to this instability. Across seven checkpoints and four benchmarks, equally small resizes cause more correctness flips when they switch configurations. Surprisingly, in over half of these cases, models answer the question incorrectly but can still read the correct answer when told what to read. Furthermore, attention interventions in LLaVA-NeXT suggest that configuration changes can weaken the use of readable information during answering. We therefore guide models using field cues and their own transcriptions. With annotation assistance, these forms of guidance together correct 97.2\% of errors with readable information. These findings show that configuration changes can affect how models use information they can still read.
\end{abstract}

%% file: sections/keywords.tex
\begin{keywords}
Vision--language models, dynamic resolution, robustness, visual evidence
\end{keywords}

%% file: sections/teaser_float.tex
\begin{figure*}[t]
\centering
\includegraphics[width=\textwidth]{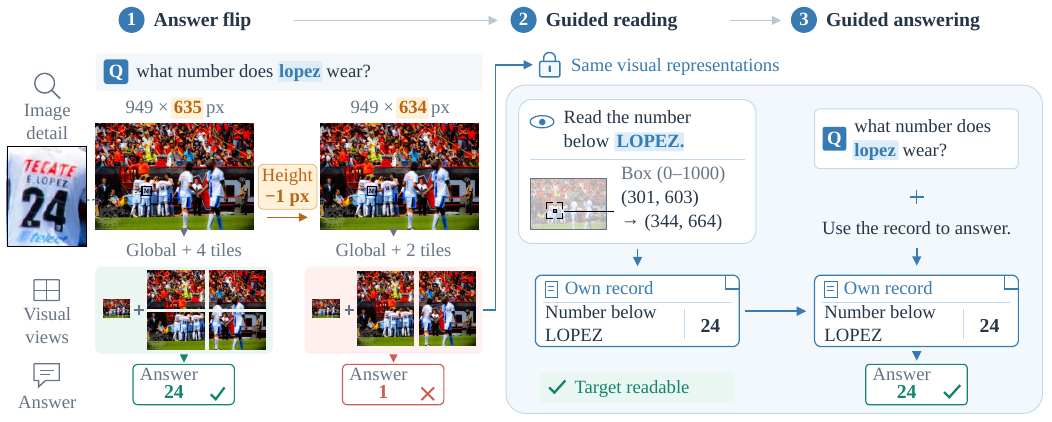}\par
\caption{A one-pixel height difference changes MiniCPM's tiling from four local tiles to two and flips the answer. Keeping the visual representations that produced the error fixed, guided reading extracts the target and the model's own field record restores the original answer. This example is from TextVQA, and all responses are generated by MiniCPM-V-4.5 using greedy decoding.}
\label{fig:teaser}
\end{figure*}

%% file: sections/introduction.tex
\section{Introduction}
\label{sec:intro}

Improving vision--language models (VLMs) for visual question answering, text recognition, and document understanding has attracted growing interest \cite{DBLP:conf/nips/LiuLWL23a,DBLP:conf/icml/0008LSH23,DBLP:conf/eccv/KimHYNPYHYHP22,DBLP:conf/icml/JiangZZY25}. A shared challenge is to preserve visual detail across image sizes within a limited computational budget \cite{DBLP:conf/cvpr/LiYLMZYSLB24,DBLP:conf/emnlp/YeHXYYXLT0ZJHLH23}. Recent models address this with dynamic resolution \cite{DBLP:conf/icml/LeeJTH0EKSCT23,DBLP:conf/eccv/GuoXYCNGCLH24,DBLP:journals/corr/abs-2511-21631,DBLP:journals/corr/abs-2509-18154}: input size determines the internal image dimensions, image tiling, and visual token grid. We call these processing choices the \textit{visual configuration}. Despite this adaptive processing, recent evaluations show that VLM performance fluctuates across resolutions and remains sensitive to spatial transformations and resampling \cite{DBLP:journals/jstsp/LiZZWTSLMLLZZ25,DBLP:conf/aaai/LiWSZHW26,DBLP:journals/corr/abs-2603-06148}.

Earlier studies in computer vision have linked similar sensitivity to how images are scaled and sampled \cite{DBLP:conf/nips/TouvronVDJ19,DBLP:conf/icml/Zhang19,DBLP:conf/cvpr/Parmar0Z22}. Recent VLM analyses provide further clues: padding can change accuracy without adding image content \cite{DBLP:conf/aaai/LiWSZHW26}, while counting failures can arise when image patterns align with the patch grid \cite{DBLP:journals/corr/abs-2607-00174}. These findings suggest that resolution sensitivity may depend on how models organize visual information. In dynamic-resolution VLMs, this organization can change abruptly with input size: a small resize may switch the token grid or image tiling. We call the input-size thresholds at which these switches occur \textit{visual configuration boundaries}. We investigate whether crossing these boundaries introduces greater answer instability than comparable changes that preserve the configuration, and present our findings through three questions:

\textbf{(RQ1)} Are VLM answers more sensitive to small resolution changes that switch the visual configuration? Across models and tasks, crossing a configuration boundary more often switches an answer between correct and incorrect than a nearby change that preserves the configuration. However, a boundary-crossing resize changes both the image pixels and the visual configuration, so their contributions must be separated. Keeping the image fixed while changing its visual configuration further shows that configuration itself contributes to this instability.

\textbf{(RQ2)} What goes wrong when a configuration switch makes the model answer incorrectly? Surprisingly, in over half of these cases, models answer the question incorrectly but can still read the correct answer when told what to read. Attention interventions in LLaVA-NeXT help explain this contrast: under the configuration that causes the error, the model relies less on the region containing the requested field (the target) when answering the original question. During \textit{guided reading}, which explicitly directs the model to read image content with field descriptions or location cues as needed, this reliance remains similar across configurations. These findings suggest that configuration changes can weaken question-guided evidence use even when the target remains readable.

\textbf{(RQ3)} Can explicit guidance help correct these errors? These findings motivate \textit{guided answering}: we use field cues and the model's own transcriptions to help it answer the original question. With annotation assistance, these forms of guidance together correct 97.2\% of errors with readable targets across four model families. Figure~\ref{fig:teaser} illustrates this progression from configuration-induced failure to successful guided reading and recovery of the correct answer.

In summary, our main contributions are threefold. \textit{(i) Configuration Boundaries.} We identify excess answer instability at configuration boundaries and isolate the contribution of visual configuration. \textit{(ii) Evidence Use.} We provide evidence that configuration changes can weaken the use of information models can still read. \textit{(iii) Guided Recovery.} We show that explicit guidance can recover original answers using information that remains readable.

%% file: sections/related_work.tex
\section{Related Work}
\label{sec:related}

\textit{Resolution sensitivity in computer vision.} Earlier work studies resolution through object scale, sampling, and training--test mismatch. SNIP \cite{DBLP:conf/cvpr/SinghD18} limits the object-scale variation seen during detector training, while FixRes \cite{DBLP:conf/nips/TouvronVDJ19} addresses differences in object scale and feature statistics between training and testing. Zhang \cite{DBLP:conf/icml/Zhang19} connects shift sensitivity to aliasing in downsampling; Clean-FID \cite{DBLP:conf/cvpr/Parmar0Z22} shows that resizing and compression can alter generative-model evaluation. These studies establish that resolution effects depend on the processing pipeline as well as the image content, motivating analysis of how models transform their visual inputs.

\textit{Dynamic-resolution visual processing.} NaViT \cite{DBLP:conf/nips/0001MDHMCSPGAOP23} supports variable resolutions and aspect ratios through sequence packing, while FlexiViT \cite{DBLP:conf/cvpr/BeyerI0CKZMTAP23} trains shared weights across patch sizes. Modern VLMs likewise adapt visual processing to input dimensions: Qwen2.5-VL \cite{DBLP:journals/corr/abs-2502-13923} varies its visual token grid, while LLaVA-UHD \cite{DBLP:conf/eccv/GuoXYCNGCLH24} and MiniCPM-V \cite{DBLP:journals/corr/abs-2509-18154} partition images into local views. These designs improve the handling of diverse inputs, but grid dimensions and tile counts change discretely. Consequently, a small resize can change both the pixels and the visual configuration. Their individual contributions to answer instability require controlled comparisons.

\textit{Resolution robustness in VLMs.} Res-Bench \cite{DBLP:conf/aaai/LiWSZHW26} measures performance across resolutions and shows that padding can change accuracy without adding image information. VLM-RobustBench \cite{DBLP:journals/corr/abs-2603-06148} identifies spatial transformations and resampling as major vulnerabilities, proposing changes in patch statistics as one explanation. Hu et al. \cite{DBLP:journals/corr/abs-2607-00174} further expose periodic counting failures when synthetic patterns align with the tokenizer's patch grid. These findings link robustness to visual processing, but do not establish whether crossing a configuration boundary adds instability beyond a nearby change that preserves the configuration. We address this gap with matched local comparisons and fixed-image configuration interventions, then examine the resulting errors through guided reading and attention interventions.

%% file: sections/boundary_table_float.tex
\begin{table*}[t]
\centering
\caption{Excess correctness-flip rate $\Delta$ (percentage points, pp). $\dagger$ indicates that the unadjusted paired 95\% confidence interval excludes zero. The per-task sample size is $n=512$ for Qwen3-8B/MiniCPM/LLaVA and $n=256$ for Qwen2.5/Qwen3-32B. For both InternVL checkpoints, $n=95$ (VQAv2), $46$ (TextVQA), $30$ (DocVQA), and $22$ (ChartQA).}
\label{tab:boundary}
{\small \input{tables/table1}}
\end{table*}

%% file: tables/table1.tex
\begingroup
\setlength{\tabcolsep}{1.25pt}
\newcommand{\metric}[2]{\makebox[3.1em][r]{#1}\makebox[0.45em][l]{\textsuperscript{#2}}}
\begin{tabular*}{\textwidth}{@{\extracolsep{\fill}}l*{8}{c}@{}}
\toprule
Task & Qwen3-VL-8B & MiniCPM-V-4.5 & LLaVA-NeXT-7B & InternVL3-8B & Qwen2.5-VL-7B & InternVL3.5-8B & Qwen3-VL-32B & Mean \\
\midrule
VQAv2 & \metric{+1.37}{} & \metric{+3.12}{\(\dagger\)} & \metric{+3.91}{\(\dagger\)} & \metric{+5.26}{} & \metric{+3.52}{} & \metric{+4.21}{\(\dagger\)} & \metric{+5.47}{\(\dagger\)} & \metric{+3.84}{} \\
TextVQA & \metric{+4.69}{\(\dagger\)} & \metric{+4.30}{\(\dagger\)} & \metric{+6.84}{\(\dagger\)} & \metric{+10.87}{} & \metric{+7.03}{\(\dagger\)} & \metric{+8.70}{\(\dagger\)} & \metric{+3.91}{\(\dagger\)} & \metric{+6.62}{} \\
DocVQA & \metric{+6.25}{\(\dagger\)} & \metric{+7.81}{\(\dagger\)} & \metric{+6.25}{\(\dagger\)} & \metric{+16.67}{} & \metric{+7.03}{\(\dagger\)} & \metric{+3.33}{} & \metric{+3.91}{\(\dagger\)} & \metric{+7.32}{} \\
ChartQA & \metric{+2.73}{\(\dagger\)} & \metric{+4.30}{\(\dagger\)} & \metric{+5.08}{\(\dagger\)} & \metric{+9.09}{} & \metric{+3.52}{} & \metric{+0.00}{} & \metric{+7.42}{\(\dagger\)} & \metric{+4.59}{} \\
\midrule
Mean & \metric{+3.76}{} & \metric{+4.88}{} & \metric{+5.52}{} & \metric{+10.47}{} & \metric{+5.27}{} & \metric{+4.06}{} & \metric{+5.18}{} & \metric{\textbf{+5.59}}{} \\
\bottomrule
\end{tabular*}
\endgroup

%% file: sections/methodology.tex
\section{Methodology}
\label{sec:method}

\subsection{Configuration Comparisons}

For an input image, the model's default preprocessing automatically selects a visual configuration based on its dimensions; we call this the image's \textit{native configuration}. When a resize crosses a configuration boundary, the images before and after resizing receive different configurations. We label these \textit{Low} and \textit{High} according to their smaller or larger number of visual tokens or tiles/views. Let $Z_c(I)$ be all visual embeddings supplied to the language decoder for image $I$ under configuration $c$. Fixing $I$ preserves the input pixels; fixing $Z_c(I)$ preserves the decoder's complete visual input.

To test whether boundaries add instability, we compare a boundary-crossing pair and a nearby control pair from each source image under the same question. Within each pair, the images differ by one pixel in height or width. The crossing images have different native configurations, whereas both control images use the same native configuration, matching one side of the crossing pair. For $N$ pairs, let $s_i^{(1)},s_i^{(2)}\in\{0,1\}$ denote answer correctness. We define the correctness-flip rate (CFR) and its boundary excess as

\begin{align}
&\mathrm{CFR}=\frac{1}{N}\sum_{i=1}^{N}|s_i^{(1)}-s_i^{(2)}|, \label{eq:cfr}\\
&\Delta=\mathrm{CFR}_{\mathrm{cross}}-\mathrm{CFR}_{\mathrm{control}}. \label{eq:boundary-excess}
\end{align}

We evaluate each crossing image under both Low and High. Changing $I$ while holding $c$ fixed tests the effect of image changes; changing $c$ while holding $I$ fixed tests the effect of configuration changes. We average the two contrasts of each type. We also compare using Low or High for both images with native processing in disagreement, accuracy, and visual-token cost.

\subsection{Probing Visual Evidence Use}

We start with crossing pairs that yield one correct and one incorrect answer. We retain only errors that can be corrected by changing the configuration while keeping the image fixed. We then test whether the target remains readable through \textit{guided reading}: using the same visual input $Z_c(I)$ that produced the wrong answer, we ask the model to read image content, adding field descriptions or location cues where needed. No reference answer values are supplied. A target is considered readable if guided reading produces its complete transcription, including required signs and units.

Attending to relevant visual regions does not guarantee a correct answer \cite{DBLP:journals/corr/abs-2510-17771}, and attention weights alone do not establish how much an output depends on a region \cite{DBLP:conf/naacl/JainW19}. We therefore use attention knockout \cite{DBLP:conf/cvpr/KaduriBD25} to test whether blocking access to target-region information changes generation. Keeping $Z_c(I)$ and the task prompt fixed, we block text attention to target-region visual tokens across all decoder layers during prompt processing and answer generation. To distinguish target dependence from disruption caused by blocking itself, we repeat the intervention on equally many non-target tokens in the same view. We compare original question answering (Q) and guided reading (R) under configurations that produce correct (G) and incorrect (B) original answers. Each task uses its own prompt, which is held fixed across the two configurations. We measure answer-content changes for Q and loss of complete target transcription for R, subtracting the non-target control effect from both rates.

We also measure changes in output probabilities. For each configuration and task, we retain the unblocked output and score that same text with and without knockout. Let $d$ denote the mean token log-probability drop after target knockout minus the drop after non-target knockout. We compute this over the full output and over key content. Here, key content refers to the first answer token that differs across configurations for Q and the first complete target span for R. The interaction
\begin{equation}
J=(d_{GQ}-d_{BQ})-(d_{GR}-d_{BR})\label{eq:interaction}
\end{equation}
compares the G-to-B change in target dependence between original answering and guided reading.

\subsection{Guided Answering}

In guided answering, we compare answer checking with three types of help (Table~\ref{tab:progressive-recovery}). \textit{Target guidance} points to annotated fields or locations \cite{DBLP:journals/corr/abs-2310-11441,DBLP:conf/cvpr/CaiLMMCPL24}. \textit{Read then answer} first asks the model to transcribe fields, describe the image, or read the target more closely, then uses that output to answer the original question, with or without an explicit target-field cue. \textit{Select then answer} uses reference field values to choose the whole record with the most correctly transcribed fields, or uses annotations to select relevant excerpts from the model's own text. The model generates these records and descriptions from the visual representations that produced the error; no reference values replace its output. In Fig.~\ref{fig:teaser}, location guidance elicits ``24'' below ``LOPEZ''; providing that record with the original question then changes the answer from ``1'' to ``24''.

Tables~\ref{tab:recovery}--\ref{tab:progressive-recovery} report cumulative recovery across these conditions. An error is counted once if any condition restores the correct answer. Table~\ref{tab:progressive-recovery} adds the four groups in order and reports only newly recovered errors at each step. This includes focused follow-ups on subsets and measures combined recovery, rather than the accuracy of a single procedure.

%% file: sections/results_and_analysis.tex
\input{sections/configuration_float}

\input{sections/common_table_float}
\section{Experiments and Analysis}
\label{sec:results}

\subsection{Experimental Setup}
\label{sec:setup}

We evaluate seven checkpoints on VQAv2 \cite{DBLP:conf/cvpr/GoyalKSBP17}, TextVQA \cite{DBLP:conf/cvpr/SinghNSJCBPR19}, DocVQA \cite{DBLP:conf/wacv/MathewKJ21}, and ChartQA \cite{DBLP:conf/acl/MasryLTJH22}: Qwen3-VL-8B \cite{DBLP:journals/corr/abs-2511-21631}, MiniCPM-V-4.5 \cite{DBLP:journals/corr/abs-2509-18154}, \href{https://llava-vl.github.io/blog/2024-01-30-llava-next/}{LLaVA-NeXT-7B}, InternVL3-8B \cite{DBLP:journals/corr/abs-2504-10479}, Qwen2.5-VL-7B \cite{DBLP:journals/corr/abs-2502-13923}, InternVL3.5-8B \cite{DBLP:journals/corr/abs-2508-18265}, and Qwen3-VL-32B \cite{DBLP:journals/corr/abs-2511-21631}. The first four serve as family representatives in Fig.~\ref{fig:configuration}, Fig.~\ref{fig:readability}, Table~\ref{tab:common}, and Tables~\ref{tab:recovery}--\ref{tab:progressive-recovery}. Low/High refers to visual-token count for Qwen and tile/view count for the other families. Field questions use TextVQA, DocVQA, ChartQA, and Visual-CoT \cite{DBLP:conf/nips/ShaoQ0SZW0024} (DocVQA, TextVQA, SROIE \cite{DBLP:conf/icdar/HuangCHBKLJ19}, and InfoVQA \cite{DBLP:conf/wacv/MathewBTKVJ22}). Decoding is greedy (96-token limit for main comparisons). Correctness uses VQA consensus scores \cite{DBLP:conf/iccv/AntolALMBZP15} $\geq$0.6 for VQAv2/TextVQA, normalized exact matching for DocVQA, and 5\% numerical tolerance for ChartQA. We use 95\% source-cluster bootstrap intervals and equally weighted task/checkpoint means.

\subsection{Instability at Configuration Boundaries}
\label{sec:boundary}

We first compare crossing and nearby non-crossing pairs using the correctness-flip rate and boundary excess defined in \eqref{eq:cfr} and \eqref{eq:boundary-excess} to test whether small resolution changes produce more correctness flips when they also switch the visual configuration. Table~\ref{tab:boundary} shows more correctness flips at crossings in nearly all model--task combinations, with a mean excess of 5.59 pp. Small resizes accompanied by configuration switches are therefore more likely to change correctness. However, both pixels and configuration change, leaving their individual contributions unresolved.

\subsection{Effects of Visual Configuration}
\label{sec:configuration}
\input{sections/readability_float}

To distinguish their contributions, we compare image changes at fixed configuration with configuration changes at fixed image. In all four families, changing configuration causes more correctness flips than changing the image, by 3.56--7.93 pp (Fig.~\ref{fig:configuration}). Complete visual configuration is therefore an important contributor to the observed instability. This finding motivates testing whether a common Low or High for both images improves stability. Both choices reduce disagreement (Table~\ref{tab:common}). High improves accuracy for MiniCPM, LLaVA, and InternVL3 at the cost of more visual tokens, whereas Low reduces token use. However, both Low and High slightly reduce accuracy for Qwen3. Thus, sharing a configuration stabilizes answers but does not guarantee higher accuracy. We therefore examine why one configuration succeeds while another fails on the same image and question.

\subsection{Question-Guided Evidence Use}
\label{sec:readability}

We first test whether the target remains readable under the configuration that causes the error. Guided reading on 300 verified configuration-induced errors recovers the complete target from unchanged visual representations in 59.3\% of cases (Fig.~\ref{fig:readability}). These results show that models can read the target when explicitly guided, although the original question fails to elicit this ability. To understand this difference, we examine how target-region visual information contributes to original question answering and guided reading.

We apply attention knockout to LLaVA-NeXT, whose visual tokens can be mapped to image regions. During original question answering, blocking target attention changes the answer less often under the configuration that produces the error, while guided reading is similarly disrupted under both configurations (Table~\ref{tab:dependence}). These results suggest that, under the configuration that causes the error, original answering relies less on the target region, while guided reading retains this reliance. We next quantify this dependence by measuring how target blocking lowers the probability of the text generated without blocking. Using \eqref{eq:interaction}, both full-output and key-content measures show a larger configuration effect during original answering than during guided reading (Table~\ref{tab:interaction}; both 95\% intervals exclude zero). This contrast suggests that explicitly telling the model what to read helps preserve the target's influence on generation across configurations. Together, these results suggest that configuration changes can weaken question-guided evidence use: readable target information contributes less reliably to the original answer through the decoder's attention pathways.

\begin{table}[ht]
\input{sections/dependence_float}

\input{sections/interaction_float}
\end{table}

\subsection{Recovery through Guided Answering}
\label{sec:guidance}

The explanation above predicts that explicit guidance should help models answer correctly using readable target information. We test this with progressively stronger guidance. Cumulative recovery rises from 18.5\% with answer checking to 82.6\% after adding target guidance and reading before answering (Table~\ref{tab:progressive-recovery}), and reaches 97.2\% (173/178) with annotation-assisted record and excerpt selection (Table~\ref{tab:recovery}). These recovered cases account for 57.7\% of all 300 configuration-induced errors. Thus, models can correctly answer the original question using information read under the configuration that caused the error, with more explicit guidance extending recovery to more cases.

\input{sections/guidance_float}

%% file: sections/configuration_float.tex
\begin{figure}[t]
\centering
\includegraphics[width=0.86\columnwidth]{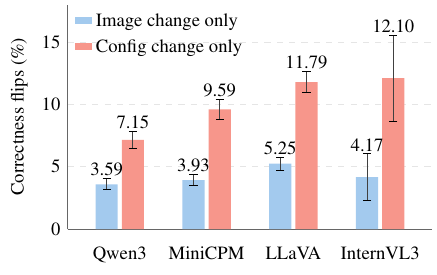}
\caption{Bars show four-task mean correctness-flip rates when changing only the image or configuration; error bars indicate 95\% confidence intervals clustered by source.}
\label{fig:configuration}
\end{figure}

%% file: sections/common_table_float.tex
\begin{table}[t]
\centering
\caption{Changes in disagreement and accuracy (pp) and visual token count (\%) under common Low/High configurations, relative to native processing.}
\label{tab:common}
{\small \input{tables/table2}}
\end{table}

%% file: tables/table2.tex
\begingroup
\setlength{\tabcolsep}{3pt}
\begin{tabular}{@{}llrrr@{}}
\toprule
Model & Config. & Flip $\Delta$ & Acc. $\Delta$ & Tokens $\Delta$ \\
\midrule
Qwen3 & Low & -3.32 & -0.34 & -3.4 \\
 & High & -4.05 & -0.02 & +3.4 \\
\specialrule{0.3pt}{1.5pt}{1.5pt}
MiniCPM & Low & -4.83 & -2.66 & -37.6 \\
 & High & -7.03 & +3.03 & +37.6 \\
\specialrule{0.3pt}{1.5pt}{1.5pt}
LLaVA & Low & -6.49 & -2.37 & -24.0 \\
 & High & -6.74 & +2.00 & +24.0 \\
\specialrule{0.3pt}{1.5pt}{1.5pt}
InternVL3 & Low & -9.03 & -0.43 & -44.2 \\
 & High & -7.38 & +0.70 & +44.2 \\
\bottomrule
\end{tabular}
\endgroup

%% file: sections/readability_float.tex
\begin{figure*}[t]
\centering
\includegraphics[width=\textwidth]{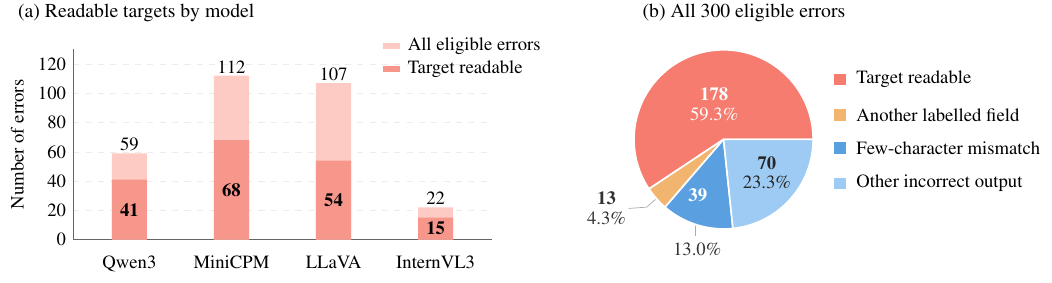}
\caption{Target readability in 300 configuration-induced errors. (a) Bars show total errors and the subset with complete target transcription through guided reading for each model. (b) The pie shows the overall distribution; remaining cases are classified by outputs under a common reading prompt.}
\label{fig:readability}
\end{figure*}

%% file: sections/dependence_float.tex
\centering
\caption{Target-attention knockout effects in LLaVA-NeXT, after subtracting non-target effects (\%).}
\label{tab:dependence}
{\small
\begin{tabular}{@{}lrr@{}}
\toprule
Configuration & Original Q & Guided reading R \\
\midrule
Correct & \textbf{82.1\%} & 71.8\% \\
Incorrect & \textbf{38.5\%} & 74.4\% \\
\bottomrule
\end{tabular}
}
\par\addvspace{\floatsep}

%% file: sections/interaction_float.tex
\centering
\caption{Interaction between configuration and task in control-adjusted log-probability drops, averaged over tokens (nats). Confidence intervals use 5,000 bootstrap resamples of sources.}
\label{tab:interaction}
{\small
\begin{tabular}{@{}lrr@{}}
\toprule
Measure & Mean interaction & 95\% CI \\
\midrule
Full output & +0.451 & [0.260, 0.643] \\
Key content & +1.596 & [0.843, 2.411] \\
\bottomrule
\end{tabular}
}

%% file: sections/guidance_float.tex
% Keep the recovery table intact without forcing an extra column or page.
\par\addvspace{\intextsep}
\noindent\begin{minipage}{\columnwidth}
\centering
\captionof{table}{Cumulative guided-answering recovery by model.}
\label{tab:recovery}
{\small \input{tables/table5}}
\end{minipage}
\par\addvspace{\intextsep}

\noindent\begin{minipage}{\columnwidth}
\centering
\captionof{table}{Cumulative recovery as guidance groups are added ($n=178$). $\dagger$ includes reference-assisted selection.}
\label{tab:progressive-recovery}
{\small \input{tables/table6}}
\end{minipage}
\par\addvspace{\intextsep}

%% file: tables/table5.tex
\begingroup
\setlength{\tabcolsep}{4pt}
\begin{tabular}{@{}lrrr@{}}
\toprule
Model & Readable & Recovered & Rate \\
\midrule
Qwen3 & 41 & 39 & 95.1\% \\
MiniCPM & 68 & 67 & 98.5\% \\
LLaVA & 54 & 52 & 96.3\% \\
InternVL3 & 15 & 15 & 100.0\% \\
\midrule
\textbf{Guided answering overall} & \textbf{178} & \textbf{173} & \textbf{97.2\%} \\
\bottomrule
\end{tabular}
\endgroup

%% file: tables/table6.tex
% Author-approved cumulative groups; each recovered error counted once.
\begingroup
\setlength{\tabcolsep}{3pt}
\begin{tabular*}{\columnwidth}{@{\extracolsep{\fill}}lrr@{}}
\toprule
Guidance added & New & Cumulative \\
\midrule
Answer checking & +33 & 33/178 (18.5\%) \\
Target guidance & +32 & 65/178 (36.5\%) \\
Read then answer & +82 & 147/178 (82.6\%) \\
Select then answer$^\dagger$ & +26 & \textbf{173/178 (97.2\%)} \\
\bottomrule
\end{tabular*}
\endgroup

%% file: sections/conclusion.tex
\section{Conclusion}
\label{sec:conclusion}

We identify visual configuration boundaries as points of increased answer instability and establish the contribution of complete configurations. Guided reading and LLaVA-NeXT attention interventions suggest that configuration changes can weaken question-guided evidence use even when the target remains readable. Guided answering shows that many original answers remain recoverable with explicit help, although reliable automatic recovery remains unresolved. Extending this analysis beyond local boundaries could clarify how visual configurations shape evidence use across resolutions, providing a broader understanding of VLM resolution robustness.

%% file: sections/ethics.tex
\section*{Compliance With Ethical Standards}

This study uses only existing publicly available datasets and involves no new human or animal studies.

%% file: main.bbl
\begin{thebibliography}{10}

\bibitem{DBLP:conf/nips/LiuLWL23a}
H.~Liu, C.~Li, Q.~Wu, and Y.~J. Lee,
\newblock ``{Visual Instruction Tuning},''
\newblock in {\em NeurIPS}, 2023.

\bibitem{DBLP:conf/icml/0008LSH23}
J.~Li, D.~Li, S.~Savarese, and S.~C.~H. Hoi,
\newblock ``{BLIP-2: Bootstrapping Language-Image Pre-training with Frozen
  Image Encoders and Large Language Models},''
\newblock in {\em ICML}, 2023, pp. 19730--19742.

\bibitem{DBLP:conf/eccv/KimHYNPYHYHP22}
G.~Kim, et~al.,
\newblock ``{OCR-Free Document Understanding Transformer},''
\newblock in {\em ECCV (28)}, 2022, pp. 498--517.

\bibitem{DBLP:conf/icml/JiangZZY25}
J.-P. Jiang, T.~Zhou, D.-C. Zhan, and H.-J. Ye,
\newblock ``{Compositional Condition Question Answering in Tabular
  Understanding},''
\newblock in {\em ICML}, 2025.

\bibitem{DBLP:conf/cvpr/LiYLMZYSLB24}
Z.~Li, et~al.,
\newblock ``{Monkey: Image Resolution and Text Label are Important Things for
  Large Multi-Modal Models},''
\newblock in {\em CVPR}, 2024, pp. 26753--26763.

\bibitem{DBLP:conf/emnlp/YeHXYYXLT0ZJHLH23}
J.~Ye, et~al.,
\newblock ``{UReader: Universal OCR-free Visually-situated Language
  Understanding with Multimodal Large Language Model},''
\newblock in {\em EMNLP (Findings)}, 2023, pp. 2841--2858.

\bibitem{DBLP:conf/icml/LeeJTH0EKSCT23}
K.~Lee, et~al.,
\newblock ``{Pix2Struct: Screenshot Parsing as Pretraining for Visual Language
  Understanding},''
\newblock in {\em ICML}, 2023, pp. 18893--18912.

\bibitem{DBLP:conf/eccv/GuoXYCNGCLH24}
Z.~Guo, et~al.,
\newblock ``{LLaVA-UHD: An LMM Perceiving Any Aspect Ratio and High-Resolution
  Images},''
\newblock in {\em ECCV (83)}, 2025, pp. 390--406.

\bibitem{DBLP:journals/corr/abs-2511-21631}
{Qwen Team},
\newblock ``{Qwen3-VL Technical Report},''
\newblock {\em CoRR}, vol. abs/2511.21631, 2025.

\bibitem{DBLP:journals/corr/abs-2509-18154}
T.~Yu, et~al.,
\newblock ``{MiniCPM-V 4.5: Cooking Efficient MLLMs via Architecture, Data, and
  Training Recipe},''
\newblock {\em CoRR}, vol. abs/2509.18154, 2025.

\bibitem{DBLP:journals/jstsp/LiZZWTSLMLLZZ25}
C.~Li, et~al.,
\newblock ``{R-Bench: Are Your Large Multimodal Model Robust to Real-World
  Corruptions?},''
\newblock {\em IEEE J. Sel. Top. Signal Process.}, vol. 19, no. 7, pp.
  1349--1361, 2025.

\bibitem{DBLP:conf/aaai/LiWSZHW26}
C.~Li, Z.~Wang, Y.~Sheng, X.~Zhu, Y.~Hao, and X.~Wang,
\newblock ``{Res-Bench: Benchmarking the Robustness of Multimodal Large
  Language Models to Dynamic Resolution Input},''
\newblock in {\em AAAI}, 2026, pp. 31545--31553.

\bibitem{DBLP:journals/corr/abs-2603-06148}
R.~Saxena, A.~Suglia, and P.~Minervini,
\newblock ``{VLM-RobustBench: A Comprehensive Benchmark for Robustness of
  Vision-Language Models},''
\newblock {\em CoRR}, vol. abs/2603.06148, 2026.

\bibitem{DBLP:conf/nips/TouvronVDJ19}
H.~Touvron, A.~Vedaldi, M.~Douze, and H.~J{\'e}gou,
\newblock ``{Fixing the train-test resolution discrepancy},''
\newblock in {\em NeurIPS}, 2019, pp. 8250--8260.

\bibitem{DBLP:conf/icml/Zhang19}
R.~Zhang,
\newblock ``{Making Convolutional Networks Shift-Invariant Again},''
\newblock in {\em ICML}, 2019, pp. 7324--7334.

\bibitem{DBLP:conf/cvpr/Parmar0Z22}
G.~Parmar, R.~Zhang, and J.-Y. Zhu,
\newblock ``{On Aliased Resizing and Surprising Subtleties in GAN
  Evaluation},''
\newblock in {\em CVPR}, 2022, pp. 11400--11410.

\bibitem{DBLP:journals/corr/abs-2607-00174}
K.~Hu, A.~Bharadwaj, W.~Yu, and M.~Fredrikson,
\newblock ``{Steal the Patch Size: Adversarially Manipulate Vision-Language
  Models},''
\newblock {\em CoRR}, vol. abs/2607.00174, 2026.

\bibitem{DBLP:conf/cvpr/SinghD18}
B.~Singh and L.~S. Davis,
\newblock ``{An Analysis of Scale Invariance in Object Detection -- SNIP},''
\newblock in {\em CVPR}, 2018, pp. 3578--3587.

\bibitem{DBLP:conf/nips/0001MDHMCSPGAOP23}
M.~Dehghani, et~al.,
\newblock ``{Patch n' Pack: NaViT, a Vision Transformer for any Aspect Ratio
  and Resolution},''
\newblock in {\em NeurIPS}, 2023.

\bibitem{DBLP:conf/cvpr/BeyerI0CKZMTAP23}
L.~Beyer, et~al.,
\newblock ``{FlexiViT: One Model for All Patch Sizes},''
\newblock in {\em CVPR}, 2023, pp. 14496--14506.

\bibitem{DBLP:journals/corr/abs-2502-13923}
S.~Bai, et~al.,
\newblock ``{Qwen2.5-VL Technical Report},''
\newblock {\em CoRR}, vol. abs/2502.13923, 2025.

\bibitem{DBLP:journals/corr/abs-2510-17771}
Z.~Liu, et~al.,
\newblock ``{Seeing but Not Believing: Probing the Disconnect Between Visual
  Attention and Answer Correctness in VLMs},''
\newblock {\em CoRR}, vol. abs/2510.17771, 2025.

\bibitem{DBLP:conf/naacl/JainW19}
S.~Jain and B.~C. Wallace,
\newblock ``{Attention is not Explanation},''
\newblock in {\em NAACL-HLT (1)}, 2019, pp. 3543--3556.

\bibitem{DBLP:conf/cvpr/KaduriBD25}
O.~Kaduri, S.~Bagon, and T.~Dekel,
\newblock ``{What's in the Image? A Deep-Dive into the Vision of Vision
  Language Models},''
\newblock in {\em CVPR}, 2025, pp. 14549--14558.

\bibitem{DBLP:journals/corr/abs-2310-11441}
J.~Yang, H.~Zhang, F.~Li, X.~Zou, C.~Li, and J.~Gao,
\newblock ``{Set-of-Mark Prompting Unleashes Extraordinary Visual Grounding in
  GPT-4V},''
\newblock {\em CoRR}, vol. abs/2310.11441, 2023.

\bibitem{DBLP:conf/cvpr/CaiLMMCPL24}
M.~Cai, et~al.,
\newblock ``{ViP-LLaVA: Making Large Multimodal Models Understand Arbitrary
  Visual Prompts},''
\newblock in {\em CVPR}, 2024, pp. 12914--12923.

\bibitem{DBLP:conf/cvpr/GoyalKSBP17}
Y.~Goyal, T.~Khot, D.~Summers-Stay, D.~Batra, and D.~Parikh,
\newblock ``{Making the V in VQA Matter: Elevating the Role of Image
  Understanding in Visual Question Answering},''
\newblock in {\em CVPR}, 2017, pp. 6325--6334.

\bibitem{DBLP:conf/cvpr/SinghNSJCBPR19}
A.~Singh, et~al.,
\newblock ``{Towards VQA Models That Can Read},''
\newblock in {\em CVPR}, 2019, pp. 8317--8326.

\bibitem{DBLP:conf/wacv/MathewKJ21}
M.~Mathew, D.~Karatzas, and C.~V. Jawahar,
\newblock ``{DocVQA: A Dataset for VQA on Document Images},''
\newblock in {\em WACV}, 2021, pp. 2199--2208.

\bibitem{DBLP:conf/acl/MasryLTJH22}
A.~Masry, D.~X. Long, J.~Q. Tan, S.~R. Joty, and E.~Hoque,
\newblock ``{ChartQA: A Benchmark for Question Answering about Charts with
  Visual and Logical Reasoning},''
\newblock in {\em ACL (Findings)}, 2022, pp. 2263--2279.

\bibitem{DBLP:journals/corr/abs-2504-10479}
J.~Zhu, et~al.,
\newblock ``{InternVL3: Exploring Advanced Training and Test-Time Recipes for
  Open-Source Multimodal Models},''
\newblock {\em CoRR}, vol. abs/2504.10479, 2025.

\bibitem{DBLP:journals/corr/abs-2508-18265}
W.~Wang, et~al.,
\newblock ``{InternVL3.5: Advancing Open-Source Multimodal Models in
  Versatility, Reasoning, and Efficiency},''
\newblock {\em CoRR}, vol. abs/2508.18265, 2025.

\bibitem{DBLP:conf/nips/ShaoQ0SZW0024}
H.~Shao, et~al.,
\newblock ``{Visual CoT: Advancing Multi-Modal Language Models with a
  Comprehensive Dataset and Benchmark for Chain-of-Thought Reasoning},''
\newblock in {\em NeurIPS}, 2024.

\bibitem{DBLP:conf/icdar/HuangCHBKLJ19}
Z.~Huang, et~al.,
\newblock ``{ICDAR2019 Competition on Scanned Receipt OCR and Information
  Extraction},''
\newblock in {\em ICDAR}, 2019, pp. 1516--1520.

\bibitem{DBLP:conf/wacv/MathewBTKVJ22}
M.~Mathew, V.~Bagal, R.~Tito, D.~Karatzas, E.~Valveny, and C.~V. Jawahar,
\newblock ``{InfographicVQA},''
\newblock in {\em WACV}, 2022, pp. 2582--2591.

\bibitem{DBLP:conf/iccv/AntolALMBZP15}
S.~Antol, et~al.,
\newblock ``{VQA: Visual Question Answering},''
\newblock in {\em ICCV}, 2015, pp. 2425--2433.

\end{thebibliography}
